\documentclass[graybox, natbib, nosecnum]{svmult}
\bibpunct{(}{)}{;}{a}{}{,} 

\usepackage{mathptmx}       
\usepackage{helvet}         
\usepackage{courier}        
\usepackage{type1cm}        

\usepackage{makeidx}         
\usepackage{graphicx}        
\usepackage{multicol}        
\usepackage[bottom]{footmisc}
\usepackage[normalem]{ulem}	
\usepackage{hyperref}  
\usepackage{soul}   

\usepackage{newtxtext}       %
\usepackage{newtxmath}       
\usepackage{todonotes}
\usepackage{algorithm}
\usepackage{algpseudocode}
\usepackage{tabularx}
\usepackage{float}

\DeclareMathAlphabet{\pazocal}{OMS}{zplm}{m}{n}

\newcommand{\edit}[1]{{#1}}
\newcommand{\remove}[1]{}

\begin{document}

\title*{Aerial Field Robotics}
\author{Mihir Kulkarni, Brady Moon, Kostas Alexis, Sebastian Scherer}
\institute{Mihir Kulkarni \at NTNU and University of Nevada, Reno, \email{mihir.kulkarni@ntnu.no} \and Brady Moon \at Carnegie Mellon University \email{bradym@andrew.cmu.edu} \and Kostas Alexis \at NTNU, Norwegian University of Science and Technology \\\email{konstantinos.alexis@ntnu.no} \and Sebastian Scherer \at Carnegie Mellon University \email{basti@andrew.cmu.edu}} 


%
%
\maketitle
\section{Synonyms}
Aerial Field Robotics\\
Micro Aerial Vehicles in the Field\\
\edit{Unmanned Aerial Vehicles in the Field}


\vspace{-4ex}

\section{Definitions}

\vspace{-1ex}

\edit{Aerial field robotics research represents the domain of study that aims to equip unmanned aerial vehicles---and as it pertains to this chapter, specifically Micro Aerial Vehicles (MAVs)---with the ability to operate in real-life environments that present challenges to safe navigation. We present the key elements of autonomy for MAVs that are resilient to collisions and sensing degradation, while operating under constrained computational resources. We overview aspects of the state of the art, outline bottlenecks to resilient \edit{navigation} autonomy, and overview the field-readiness of MAVs. We conclude with notable contributions and discuss considerations for future research that are essential for resilience in aerial robotics.}

\vspace{-3ex}
\section{Overview}
\label{sec:intro}
\vspace{-1ex}







\edit{The state of the art in aerial robotics can accomplish impressive tasks. Yet wider use and adoption of MAVs for effective field deployment is limited by the resilience of the components of autonomy. In this chapter, we view each element of the autonomy system  under the framework of resilience and examine the latest developments, as well as open questions.}

\begin{figure*}[t]
\vspace{2ex}
\centering
    \includegraphics[width=0.99\columnwidth]{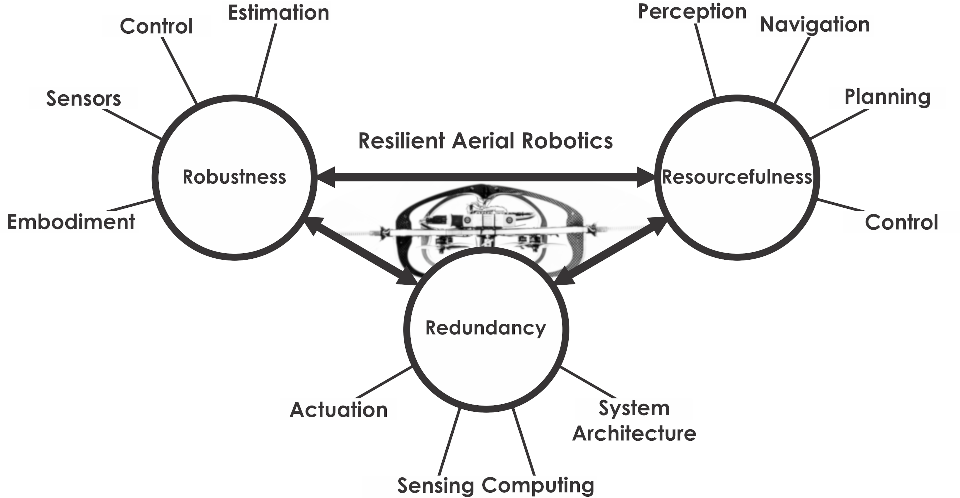}
\caption{\remove{Notional organization}\edit{Core elements} of resilient autonomy for aerial \edit{field} robotics.}\label{fig:whatisresilience}
\end{figure*}

Towards a principled understanding of progress in resilient and field-hardened aerial robotic autonomy, we define resilience motivated by analogous studies in the domain of risk analysis~\citep{howell2013global}. A system presents the virtue of resilience if it demonstrates the essential characteristics of a) robustness, b) redundancy, and c) resourcefulness as they relate to its control, perception, path planning, and decision-making. This organization is illustrated in Figure~\ref{fig:whatisresilience}.

\noindent \textbf{Robustness} incorporates the concept of reliability and refers to the ability of the robot to absorb and withstand disturbances and deteriorating effects of unpredictable situations.

%
%

\noindent \textbf{Redundancy} involves \edit{a robot's} capacity and  \edit{the presence of back-up systems to enable the maintenance of core functionality in the event of disturbances and failures by incorporating a diversity of overlapping subsystems, methods, policies, and last-resort safety schemes.}

\noindent \textbf{Resourcefulness} \edit{refers to} the ability to adapt to changes, uncertainties and crises\edit{, with the robot's inherent flexibility to deliver a certain functionality using a multitude of possible solutions. This is obtained through the combination of various subsystems that exploit their elementary capacities in different ways to reduce the likelihood of a full failure.}


%
\begin{figure}[h!]
\vspace{2ex}
\centering
    \includegraphics[width=0.8\columnwidth]{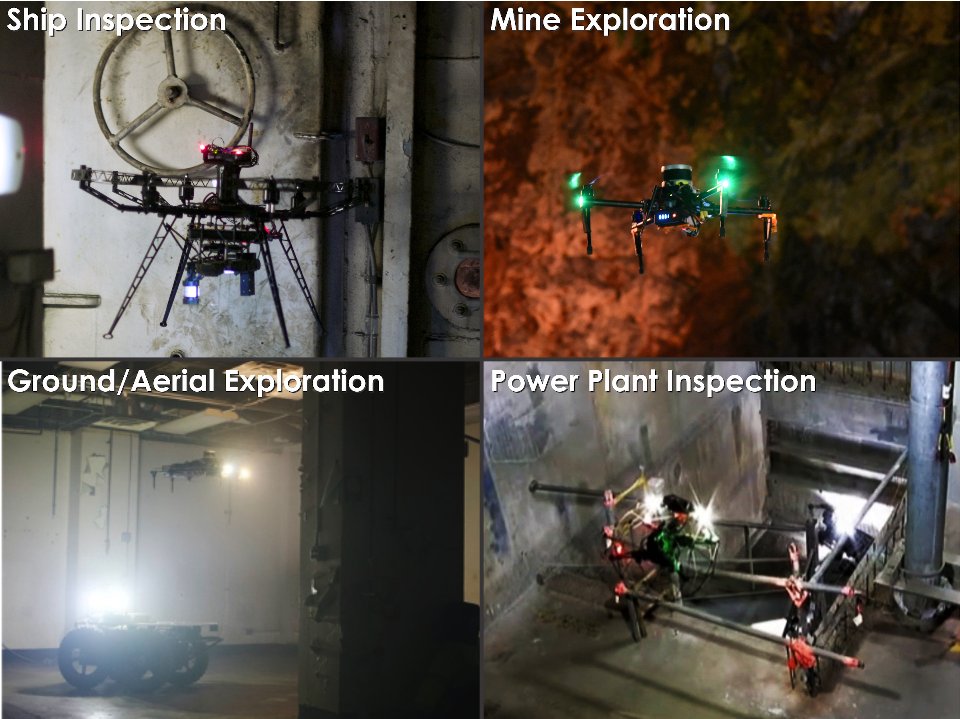}
\caption{\edit{Examples of} MAV field deployments \edit{for inspection and exploration in geometrically-complex sensor-degraded environments.} }\label{fig:introfig}
\end{figure}
%

Given this perspective on resilient autonomy, in this chapter, we focus on MAV embodiment, control, perception, and planning in cluttered \edit{perceptually-degraded} environments \edit{(Figure~\ref{fig:introfig})}. Other aerial robot configurations (e.g., fixed-wing systems) or operational \remove{environments and conditions}\edit{settings} are beyond the scope of this study. We \remove{overview}\edit{survey} the state of the art\edit{, example applications}\remove{, especially} as reflected in pioneering field results\remove{demonstrated in open competitions}, and seek to identify the \edit{open problems and} bottlenecks \edit{to} autonomy. 


\section{Key Research Findings}

\section{Resilient Embodiment}\label{sec:embodiment}


The mechatronic design of MAVs can contribute to achieving resilient operation and may mitigate limitations and risks \edit{to} the autonomy \edit{stack}. We focus on two types of aerial robot embodiment designs, namely collision-tolerant systems and shape-reconfigurable designs. \edit{This narrow scope reflects the importance of mitigating the effects of collisions.}

When tasked with navigating cluttered and narrow environments, ensuring safe traversal becomes a task of increasing complexity, as any error in localization, mapping, planning or trajectory tracking can lead to a collision. Typically, \edit{collisions are seen as catastrophic events and are reasons for mission interruption}. However, \edit{drawing inspiration from nature}, we can identify an alternative, which is to render collisions (within bounds) \edit{acceptable}\remove{as an acceptable event}. 

%
\begin{figure}[h!]
\vspace{2ex}
\centering
    \includegraphics[width=0.99\columnwidth]{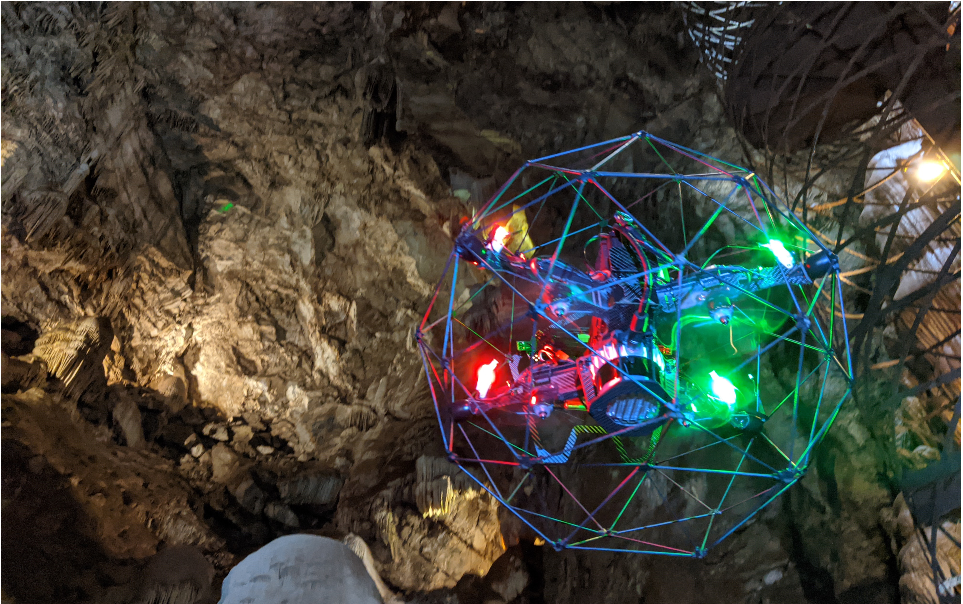}
\caption{Example of a collision-tolerant aerial robot: The Gagarin system~\citep{mbplanner} tested in subterranean environments.}\label{fig:coltolflyers}
\end{figure}
%



The community has developed MAVs that can survive collisions, with \edit{an example}\remove{examples} shown in Figure~\ref{fig:coltolflyers}. This can be achieved by developing rigid collision-tolerant outer frames that distribute the impact of a collision along the robot body to prevent localized damage~\citep{briod2014collision}. Other approaches focus on designing soft compliant structural components that cushion the robot from impact. \citep{klaptocz2013euler} utilize an Euler spring collision cage, while \citep{mintchev2017insect} realize an insect-inspired compliant design \remove{focusing on }protecting the main ``core'' with the robot electronics and sensors. Emphasizing the protection of the essential parts, \citep{salaan2019development} implement separate rotating shells around each of the quadrotor's propellers. \remove{Focusing on}\edit{For} protecting the robot when it is close to obstacles, \citep{hedayati2020pufferbot} present an actuated mechanism that drives an \remove{3D-printed }expandable scissor structure\remove{ when needed}. Motivated by origami\remove{ structures}, \citep{shu2019quadrotor} present a passive foldable airframe that acts as a protective mechanism for small multirotors. \edit{Emphasizing both compliance and the ability to re-orient in case of a crash,~\citep{zha2020collision} present tensegrity-based soft collision-tolerant MAVs.}

In addition to collision-tolerance, some works show promising results in actively or passively morphing the robot's shape to enter otherwise inaccessible regions. \citep{falanga2019foldable} develop a quadrotor with independently rotating arms around the central frame. \citep{bucki2019morphing} modify the design of an MAV replacing the rigid connections between the arms and the ``core'' with spring hinges that fold the arms downward on low thrusts. The thrust of the robot is controlled \remove{during flight }such that it allows the arms to fold, and the robot can \edit{thus} traverse narrow gaps.

\section{Resilient Control}\label{sec:control}


Control for MAVs such as quadrotors and other small rotorcraft is an essential component necessary for their stability and the efficient rejection of disturbances. \remove{This domain has been heavily researched}\edit{In this domain, a} set of high-performing and robust methods have been presented.\remove{ which fueled the success of MAVs.}

Generally, \remove{due to the difference in the time constants of the MAV position and attitude dynamics}\edit{due to the fact that the MAV attitude dynamics present much smaller time constants compared to their position dynamics}, \edit{cascade control} is commonly employed~\citep{Corke2012Multirotor}, either considering linearized or nonlinear dynamics. Different control schemes have been applied, including fixed-gains control~\citep{Bouabdallah2007FullControl}\edit{,} possibly using error functions on SO(3)/SE(3)~\citep{Lee2010Control}, model predictive control~\citep{Kamel2017LinearNonlinearMPC}, \edit{geometric $\pazocal{L}1$ control \citep{kotaru2020geometric}, $\pazocal{H}_{\infty}$ control \citep{RAFFO2008hinf},} and reinforcement learning~\citep{Hwangbo2017RL}. \edit{\citep{Powers2013} address the modelling of the systems to account for disturbances such as winds, gusts or those arising due to proximity, while \citep{McKinnon2016disturbanceEstimation} propose methods to estimate such disturbances. \citep{hentzen2019disturbance} compare the performance of methods to reject these disturbances.} \edit{Finally, systems with a higher number of actuators have been designed~\citep{kamel2018voliro}, and the intrinsic redundancy against propeller failure has been studied~\citep{mueller2014loss}}.

\section{Resilient Perception}\label{sec:perception}

Control and decision-making depend on perceiving the robot's location and understanding the world surrounding it. For MAVs it is essential that the sensors and algorithms are robust and complementary to overcome obscurants and lack of image or geometric texture\edit{,} while still being able to run efficiently onboard. The architecture of currently fielded systems relies on always-available, sufficiently accurate, low noise, and high-frequency pose estimates to enable navigation and stable flight. While some initial research has explored different paradigms to relax the requirement on estimates, current research has shown significant progress on robust state estimation and 3D mapping that can \edit{separate obscurants from thin obstacles}. \edit{To maximize resilience, robots need resourceful multi-modal approaches that can penetrate all conditions of perceptual degradation and scale across environment sizes.}

\subsubsection{Navigation in Perceptually-degraded Environments}

\begin{figure}[h!]
\centering
    \includegraphics[width=0.8\columnwidth]{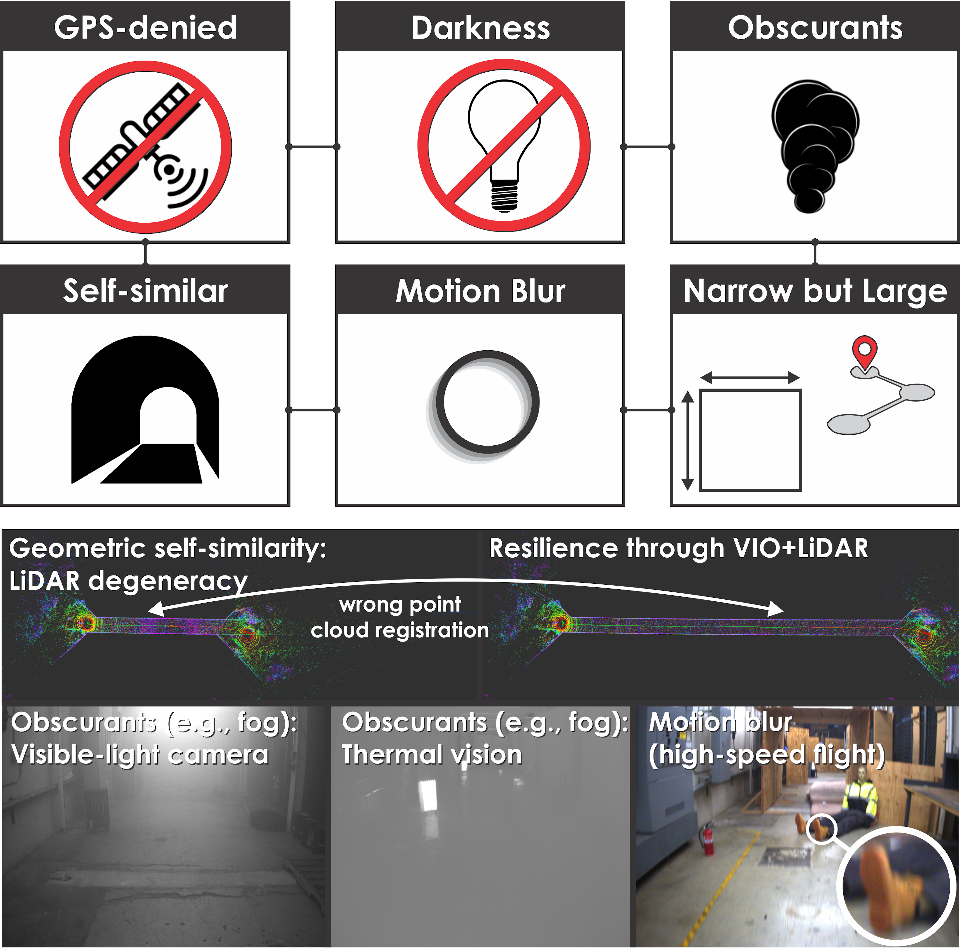}
\caption{Prominent cases of perceptual degradation and some relevant examples.}\label{fig:degradationchallenges}
\end{figure}

Researchers have contributed a set of strategies for consistent estimation and autonomous flight using sensors such as LiDARs, cameras, and thermal vision combined with Inertial Measurement Units (IMU). Monocular and stereo visual-inertial SLAM achieves impressive results~\citep{delmerico2018benchmark,Rosinol20icra-Kimera} but can get degraded in dark, textureless, or obscurant-filled settings. LiDAR-based methods present particularly robust results~\citep{zhang2014loam} but can get degenerate within self-similar geometries~\citep{zhang2016degeneracy} or environments subject to dense clouds of dust and other obscurants. To deal with \edit{certain} types of obscurants at the cost of less sharp and more noisy images, some researchers investigated LongWave InfraRed (LWIR) thermal vision-based odometry~\citep{khattak2019,delaune2019}. Furthermore, aerial robots often have to deal with varying conditions of perceptual degradation (Figure~\ref{fig:degradationchallenges}) leading to deprived sensor quality that can render sensor streams uninformative.

%
%

In response, \citep{shen2014msf} proposed the fusion of\remove{ measurements from} multiple heterogeneous sensors and used an Unscented Kalman Filter to provide smooth, globally consistent estimates.\remove{REMOVED: for position in real-time. Their system was tested in an industrial complex involving cluttered buildings, trees, and indoor environments.} Focusing on resilient localization against multiple cases of degradation, the community has focused on multi-modal sensor fusion\remove{ approaches} tailored to\remove{ the} computationally constrained MAVs. CompSLAM\remove{, which stands for Complementary multi-modal SLAM}~\citep{compslam} fuses LiDAR, visual cameras, thermal cameras, and inertial cues to \edit{resourcefully} estimate the robot pose and the map of its surroundings with resilience against the discussed \remove{conditions}\edit{cases} of degradation\remove{, to the extent that those do not appear all simultaneously}. It relies on a cascaded combination of visual/thermal-inertial fusion and LiDAR Odometry and Mapping motivated by~\citep{bloesch2015robust,zhang2014loam}. Essential to its design is the observation that if we can identify when a certain estimate, driven from vision or LiDAR\remove{ sensing}, is healthy or not, then a loosely-coupled approach can be effective. \remove{This complementary fusion principle is also found in other relevant works. }LOCUS~\citep{Palieri_2021} focuses on multi-sensor LiDAR-centric odometry and mapping with the aim to be deployable across\remove{ both} ground and flying robots. It relies on \remove{a }multi-stage scan matching\remove{ unit} equipped with a health-aware sensor integration module \remove{aiming for}\edit{towards} \remove{the }seamless fusion of additional \edit{sensors}\remove{sensing modalities}. \remove{Similarly,~}\citep{ebadi2020lamp} propose LAMP, a pipeline for exploration in perceptually-degraded environments relying on the complementary fusion of diverse modalities including LiDAR and vision. \remove{Motivated by the particular challenges of visually-degraded flight in shipboard environments, \citep{fang2017robust} propose two-layer fusion where real-time odometry results are combined with particle filter-based localization combining additional sensor cues.} \edit{Overall, while prior research was able to integrate redundant sensing modalities current research focuses on how multi-modality contributes to solution quality and on how to detect partial degradation.}

Another idea \remove{used to improve resilience }has been to explore learning-based approaches to overcome corner cases that hand-engineered algorithms cannot overcome. Examples of methods trying to increase robustness via learning are TartanVO~\citep{tartanvo2020corl}, CubeSLAM~\citep{Yang:2018eo}, and ESP-VO~\citep{doi:10.1177/0278364917734298}. \edit{Very recent efforts demonstrate the qualities of combined learning-based navigation methods and visual-inertial estimation~\citep{loquercio2021learning}.} \remove{However, these methods need large datasets to generalize across a wide range of environments.}

\remove{As important as the method itself, is understanding the constraints a sensor provides to a SLAM system to predict whether a particular area would offer enough information to localize. This is also referred to as the ``localizability'' of an environment~\citep{Zhen:2019jn}. Predicting how well a camera is able to estimate its ego-motion is hard. Therefore, many heuristics have been developed to predict sensor uncertainty. For example, uncertainty can be based on the reprojection error of inliers, the percentage of inliers, as well as other metrics.}

\subsubsection{Map Representation}\label{sec:maprepr}

Sensors in the real-world are imperfect and generate systematic and random errors due to the sensor itself, as well as degradation in the environment. These errors can be captured in a sensor model and are typically filtered using \edit{Bayesian} filtering. These sensor models are then used to update representations for perception. While continuous representations such as Gaussian processes~\citep{tabib2019real} have been used, regular volumetric discretization in evidence/occupancy grids is a fast and often sufficiently robust method. Octomap~\citep{hornung2013octomap} is a popular framework that uses octrees to store occupancy information of the environment. However, octrees have complexity of $O(logn)$ for insertion and look-up, where $n$ is the tree depth, rendering building and querying the map in large environments inefficient. Recent works including~\citep{oleyknikova2017voxblox} utilize voxel hashing for fast \edit{$O(1)$} lookups of voxel information from 3D coordinates. This allows incrementally constructing Euclidean Signed Distance Fields maps from \remove{the }Truncated Signed Distance Fields (TSDF).\edit{~\citep{Whelan-2012-7552} used a TSDF with a cyclical buffer, while} OpenVDB~\citep{museth2013vdb} \remove{implements}\edit{uses} a hierarchical data structure offering effectively infinite $3\textrm{D}$ index space, exploits spatial coherency of time-varying data, imposes no topology restrictions, and supports $O(1)$ random access. \edit{Notably, in the above we have considered the common practice of assuming that the state estimate mean is correct. New research considers the role of noise and non globally-consistent estimation~\citep{Doherty2019,cieslewski2019exploration}.} 

\subsubsection{Difficult to Detect \& Avoid Obstacles}

When trying to perceive objects within the surrounding environment, certain factors can render objects hard-to-detect including their size, shape, material type, or coating \edit{which can lead to inaccurate maps and potential collisions.} 

\remove{Poles and wires are hard-to-detect because of how thin they are. } The small cross-section of poles and wires gives a weak signal for both cameras and radars. \citep{madann2017} use Convolutional Neural Networks (CNN) to detect wires in monocular images and used synthetic data to train the model. \remove{Later work by~}\citep{stambler2019} also \remove{uses}\edit{use} a CNN and \remove{includes }methods for tracking the wires in $3\textrm{D}$. \edit{Trained purely on real-world datasets, the model outperformed the previous methods on precision and computation time.} \remove{, while still highlighting the challenges and shortcomings of detecting wires only with a monocular camera.}

Detecting reflective and transparent objects is also a challenge. \citep{dubey2018} use a CNN to segment thin obstacles (i.e., wire pixels)\remove{ with a \remove{binocular }stereo pair to detect other generic obstacles}. Another method of detecting transparent objects uses a laser rangefinder and looks for the spike in reflected light as the incident angle approaches the surface normal of the transparent object~\citep{WANG201797}\edit{, relying} on the laser being able to hit the object near its surface normal.

\section{Resilient Planning}
\label{sec:planning}

The challenge of resilient aerial robotic autonomy in the field depends upon their capacity to \edit{resourcefully} plan actions that \remove{efficiently}\edit{robustly} optimize the robot's mission objectives (``extrinsic goals'') and simultaneously ensure the health and safety of the robot (``intrinsic motivations'').

\subsubsection{Problem Definition}

The overall problem considered relates to that of \textit{iteratively} identifying admissible paths that optimize a given set of objectives that may represent extrinsic mission goals (e.g., to search an area) or intrinsic motivations (e.g., to reduce localization uncertainty) given a sequence of online acquired observations\remove{"and possibly an incrementally reconstructed map"} of the environment. An admissible path is one that ensures the safety of the vehicle (e.g., guarantees collision-avoidance), and accounts for limitations of the platform.

\subsubsection{Path Planning for Aerial Field Robotics}

Below we present selected literature and path planning architectures for a class of problems involving diverse objectives.  \remove{Figure outlines the key components of a baseline path planning architecture. andFigure 5, Path Planning Kernel Figure}

%
%

\subsubsection{Path Search}

The prime approach to enable path planning for aerial robots involves the (online) reconstruction of a map of the scene. Provided a map, the planning architecture employs a policy for searching admissible paths \edit{that optimizes specified objectives for the robot.}

\noindent \textbf{Planning to known destinations:} In the simplest case, the goal is to arrive at a pre-defined location by avoiding \remove{any }obstacles\remove{that are} on the way.\remove{To that end, the} The prevailing methods in the literature involve sampling-based path planners that sample in the robot's configuration space~\citep{rrtstar} or the\remove{robot's} control space~\citep{liu2018motion}. Focusing on computational speed, SPARTAN~\citep{cover2013sparse} creates a sparse graph from vertices that lie on the surface of the collision \edit{space, where the optimal path between any two states is shown to be }approximately tangential to the surface of the (disjoint) collision sub-spaces. \edit{Beyond such techniques, recent efforts on geometric trajectory optimization have shown exciting results on agile navigation given a priori map knowledge~\citep{wang2021geometrically}.}

\noindent \textbf{Planning to optimize extrinsic mission objectives:} Beyond the case where the goal destination is known, planning for aerial robotics has extended to include strategies that satisfy more complex mission objectives. \edit{Exploration of unknown environments is an example of an extrinsic objective of particular interest, especially as unknown environments may present unforeseen obstacles, involve cluttered regions, hazardous operating conditions, multi-storeyed and multi-branching topologies.} To address the challenge, researchers have investigated a range of approaches \remove{based on}\edit{including~~}frontier-based exploration, sampling-based methods, as well as learning-based techniques.



In the first case, the general approach includes the identification of frontiers and respective configurations towards which the robot is guided given a method for collision-free motion planning. \citep{nuske2015autonomous} present a goal-point extraction strategy for river exploration that maximizes the information gained in the MAV mission by introducing multivariate cost maps. \edit{Fast frontier-based exploration is achieved in~\citep{dai2020fast} exploiting implicit frontier voxels grouping and planning of next viewpoints based on map entropy and travel time derivations.}

\remove{Newtonian particle-based exploration}

\edit{Sampling-based} path search is investigated to identify admissible paths which are also evaluated in terms of information gain. \citep{nbvp} proposed the use of rapidly-exploring random trees \edit{RRTs} to search for collision-free paths and assess a volumetric exploration gain along each of the tree branches. This method,\remove{called Receding Horizon Next-Best-View-Planner (NBVP)} extracts the best path, and commands the robot to go to its first vertex. This procedure is then repeated in a receding horizon fashion. \citep{gbplannerjfr} focused on large-scale and possibly narrow subterranean environments\edit{, and proposed a bifurcated architecture of dense local and sparse global graph search}. Motivated by this method but focusing on \remove{agile exploration, \citep{mbplanner} employ motion primitives at its local planner}\edit{combined exploration and coverage, \citep{habplanner} employs a multi-hypothesis approach for multi-objective optimization}. \remove{Observing the fact that}\edit{Since} sampling-based methods \edit{relying on information gain formulations present the pitfall of being stuck in a local region and do not reach global coverage}, \citep{schmid2020efficient} proposed \edit{an RRT$^\star$-inspired planner that expands a single tree of candidate trajectories and uses a novel reconstruction gain and cost formulation that allows to perform global coverage and minimize path cost in the global context using a single objective function.}

Additionally, \edit{learning-based} path planning policies \remove{that rely on learning-based techniques}have emerged and encompass the possibility to outperform the state of the art. Focusing specifically on subterranean environments,~\citep{reinhart2020learning} propose a supervised learning agent \edit{for} autonomous exploration \remove{without assuming access to an online reconstructed map of the environment}\edit{using only point cloud data}. \citep{jung2018perception} contribute a perception, guidance and navigation system for indoor autonomous drone racing\remove{ using deep learning}. Exploiting the ability to learn visuomotor policies, \citep{bonatti2019learning} present\remove{drone racing} navigation \edit{for drone racing} using cross-modal representations. Focusing on \edit{following cluttered forest trails} with an MAV equipped with a single camera, \citep{giusti2015machine} contribute a deep neural network-based image classifier to output the \edit{trail's} primary direction. \citep{loquercio2018dronet} contribute DroNet, a CNN \remove{based}\edit{trained} on driving/bicycle data \remove{but }\edit{tuned} to guide an MAV flying inside cities\remove{ across different altitudes}.

\noindent \textbf{Co-optimization of intrinsic objectives:} By ``intrinsic motivations'' we refer to all those objectives that do not relate directly to the specified mission but relate to the robot's safety and thus its ability to execute its mission reliably. The case of belief-space planning accounting for state estimation uncertainty \textit{(e.g., due to perceptual degradation)} is of particular interest. \citep{achtelik2014motion} propose to employ Rapidly exploring Random Belief Trees (RRBTs) to identify paths that lead to a waypoint, while inherently avoiding motion in unobservable modes. Focusing on autonomous exploration, \citep{papachristos2017uncertainty} propose a two-step scheme to first \edit{identify an exploration path and then derive refined} paths to the first viewpoint minimizing the anticipated localization uncertainty. \edit{Perception-aware model predictive control~\citep{falanga2018pampc} unifies control and planning to optimize perception objectives to ensure reliable sensing.} These works share similarities with a broader contribution on perception-aware planning~\citep{costante2016perception}. \edit{To enable smooth and fast traversal of tightly constrained environments, trajectory generation and optimization algorithms attempt to minimize kinodynamic objectives~\citep{mellinger2011snap}, allowing fast replanning~\cite{zhou2019fast}, and risk-aware trajectory refinement to avoid unseen obstacles~\citep{zhou2020raptor}.}

\section{Examples of Application} 
\label{sec:fieldexperience}
\edit{With an ever-increasing need to deploy reliable and robust robots in the field, there has been a strong push for developing autonomous aerial robots for challenging environments. We present a selected set among notable achievements that have been realized in this domain and further discuss challenges and bottlenecks to resilient autonomy.} We first focus on four selected open competitions that have helped to accelerate research towards resilient and field-hardened aerial robotic autonomy, namely a) the DARPA \edit{Subterranean (SubT)} Challenge, b) the DARPA Fast Lightweight Autonomy (FLA), c) the Mohamed Bin Zayed International Robotics Challenge(s) (MBZIRC), and d) the Lockheed Martin AlphaPilot Innovation Challenge. The results from such competitions are significant as teams do not have the safety net of repeating an experiment multiple times \edit{or in a controlled manner}. A set of notable MAV demonstrations, the platforms and hardware used, and significant milestones are detailed in (Table~\ref{tab:ecB}).

\begin{table*}[t]
    \begin{tabular}{ | m{1.5cm} | m{1.7cm} | m{2.3cm} | m{5.5cm} | } 
         \hline
         \textbf{TEAM} &  \textbf{PLATFORM} & \textbf{SENSING} & \textbf{MILESTONE IN MAVs} \\
         \hline
         \hline
         \multicolumn{4}{|c|}{\textbf{DARPA Subterranean Challenge}} \\
         \hline
         Explorer & Collision tolerant quadrotor & 3D LiDAR, stereo cameras & Marsupial deployment of collision tolerant MAV. Use of topological maps to direct exploration steps \citep{Scherer:2021}.\\
         \hline
         CSIRO & \edit{Aeronavics NAVI} &  3D rotating LiDAR payload, gimbal mounted camera & Marsupial deployment of MAV. Navigation through vertical shaft. Explore and Sync strategy for periodic network connectivity \citep{williams2020csiro}. \\
         \hline
         CERBERUS & DJI-M100, Collision tolerant quadrotors & 3D LiDAR, camera, \edit{thermal vision}, IMU & Autonomous negotiation of staircases. Collision tolerant MAV. Path refinement for safer traversal \citep{gbplannerjfr}. \\
         \hline
         CTU-CRAS-NORLAB & \edit{Custom Quadrotor} & 3D LiDAR, stereo and RGB cameras & Autonomous operation without operator interference. \edit{$25\textrm{min}$ flight time}\citep{ctu2019subt}. \\
         \hline
         NCTU & Custom Blimp & Stereo camera, IR sensor & Lightweight, collision-tolerant platform. Upto 90 min flight time \citep{huang2019duckiefloat}. \\
         \hline
         \hline
         \multicolumn{4}{|c|}{\textbf{DARPA Fast Lightweight Autonomy Program (FLA)}} \\
         \hline
         SSCI- AeroVironment & DJI-F450 & Camera, IMU & Steering field controller for obstacle avoidance. Visual-inertial navigation system using monocular camera. Speeds upto $19.0\textrm{m/s}$ \citep{escobar2018r}. \\
         \hline
         UPenn & DJI-F450 & 2D nodding LiDAR, stereo cameras, IMU & Autonomous navigation through indoor stairwells. Novel stereo-VIO\remove{ run onboard}.  Speeds upto $18\textrm{m/s}$ \citep{mohta2018fla,mohta2018fast}. \\
         \hline
         MIT-DRAPER & DJI-F450 & Stereo cameras & Aggressive flight in urban environments. Use of minimum-jerk trajectories. Speeds upto $9.4\textrm{m/s}$ \citep{ryll2019efficient}. \\
         \hline
         \hline
         \multicolumn{4}{|c|}{\textbf{AlphaPilot}} \\
         \hline
         UZH-RPG & Standardized quadrotor & Stereo cameras, IMU, laser rangefinder & Simultaneous detection of multiple gates for drift compensation. Speeds upto $8.0\textrm{m/s}$ \citep{foehn2020alphapilot}. \\
         \hline
         TU Delft-MAVlab & Custom-$72$g quadrotor & JeVois smart camera & Lightweight snake-gate detection. Visual model-predictive localisation \citep{li2020visual}. \\
         \hline
\end{tabular}
    \caption{Selected competitions and events.}
    \label{tab:ecB}
\end{table*}

The DARPA \edit{SubT} Challenge \edit{called} for teams to deploy a robotic system-of-systems capable of autonomously exploring diverse underground environments (e.g., mines, metropolitan subterranean infrastructure, caves), detecting objects of interest and correctly reporting their location, alongside broadly mapping such environments. \edit{The Final Event of the competition took place in 2021 and required teams to deploy robots to navigate a large, multi-branching environment, involving both narrow and wide cross-sections, cluttered spaces, and dynamic obstacles in GPS-denied and visually-degraded conditions. Team CERBERUS~\citep{tranzatto2021cerberus} won first place and - like other teams~\citep{agha2021nebula} - developed custom collision-tolerant flying platforms capable of autonomous navigation through such adverse environments to perform exploration and search missions. Collision-tolerance and sensing multi-modality were essential for the deployability of MAVs underground. Multiple teams used ground robots to carry and deploy the MAVs inside the environment in a ``marsupial'' fashion. While almost all the participating teams used aerial robots, they did not evolve to become their primary platform of choice, except for Team Explorer which scored one-third of their points with aerial robots, highlighting the limitations in endurance and computational capabilities.}


The DARPA FLA program aimed to explore non-traditional perception and autonomy to enable high-speed MAV navigation in cluttered environments. The program specifically aimed for autonomous, GPS-denied, \edit{fast} navigation. \edit{The challenges included flying at increased speeds between multi-story buildings, through tight alleyways and narrow openings, into buildings, searching rooms and creating $3\textrm{D}$ maps of the interior. The challenge was particularly interesting as it led to the development of robust and lightweight estimation, planning, and control approaches for computationally constrained platforms.} A stereo Multi-State Constraint Kalman Filter was used \edit{by the UPenn's team} to provide computationally efficient and robust odometry estimation~\citep{sun2018smsckf} achieving fast flight (of $17.5\textrm{m/s}$). The MIT-DRAPER team used a closed-form trajectory generation to obtain minimum-jerk trajectories supporting accurate visual-inertial odometry, with the robot speeds planned based on obstacle densities ~\citep{ryll2019efficient}. The SSCI-AeroVironment team used a monocular camera to navigate the cluttered environment at high speeds~\citep{escobar2018r}, \edit{using the expansion rate of objects in the field of view, a steering-field controller followed an instantaneous steering direction that minimized the risk of collision.}




The AlphaPilot competition, organized by Lockheed Martin and the Drone Racing League, challenged teams to develop AI-enabled frameworks that could navigate a fully autonomous aerial robot through complex, multi-dimensional racing courses \edit{without any pre-programming or human intervention.} \edit{The competition focused on pushing the capabilities of vision-based navigation at high-speeds with limited computation available.} The final race\remove{ event} took place in \remove{Austin, }Texas with MAVLab winning the competition, while the second team - UZH Robotics \& Perception Group - was only $3$ seconds slower. The MAVLab team deployed a learning-based visual-odometry system~\citep{li2020visual} which allowed for a high update rate and thus high-speed flight. Despite this outstanding result, the capabilities \remove{demonstrated by the}\edit{of} autonomous \remove{systems}\edit{MAVs} are not yet on par with expert human pilots \edit{and there is a push to surpass the high speeds and agility demonstrated by the humans~\citep{wang2021geometrically}.}


The MBZIRC is a series of competition events. The MBZIRC2020 consisted of three challenges and a triathlon-like Grand Challenge. \remove{In all challenges, MAVs were essential. }In the first challenge, a team of aerial robots was tasked to \edit{track and interact with a set of objects autonomously}\remove{autonomously track and interact with a set of objects} (e.g., an intruder MAV) following $3\textrm{D}$ trajectories in an indoor setting. In the second challenge, a team of aerial and ground robots was tasked to collaborate to locate, pick, transport and assemble different types of brick-shaped objects to build predefined structures outdoors. In the third challenge, a team of aerial and ground robots \remove{were}\edit{was} tasked to autonomously extinguish a series of simulated fires in a high-rise building. \edit{In the Grand Challenge, the mission task required a team of $3$ MAVs and $1$ ground robot to compete in a triathlon combining the three previous events. The overall \edit{need} for robot team coordination, accurate navigation and precise control pushed the frontier of resilience in multi-robot systems and MAVs.} \remove{More than $30$ teams were selected to participate. }The \edit{top-scoring} team in the Grand Challenge consisted of members from Czech Technical University in Prague, University of Pennsylvania, and New York University. \remove{The competition raised the barrier for multi-robot systems executing complex tasks with speed and efficiency.}



Beyond the above results demonstrated in the context of the highlighted competitions, the community has reached critical milestones in the domains of robust perception, planning and mapping for long-term deployment of MAVs. Below we discuss achievements in the following directions: a) Autonomy in Extreme Environments, b) High-speed Navigation, c) Multi-Robot Teaming and d) Unconventional Applications of Aerial Robots.

\noindent\textbf{Autonomy in Extreme Environments:} \edit{Focusing on field tests conducted within extreme indoor or highly cluttered outdoor settings,~\citep{goel2021rapid} details a multi-robot distributed mapping approach for rapid exploration in reduced bandwidth scenarios using a team of aerial robots.~\citep{yang2021graph} demonstrates topological exploration in large-scale $3\textrm{D}$ underground settings. Aiming to facilitate \remove{an intelligent response to}\edit{efficiency against} topologically complex underground environments,~\citep{mansouri2019visual} proposes visual subterranean junction recognition for MAVs using CNNs.}

\noindent\textbf{High-speed Navigation:} \edit{At higher flight speeds, the simplistic models become increasingly inaccurate as the aerodynamic drag becomes significant and must be considered into the quadrotor's dynamic model. \citep{marc2019} demonstrate accurate trajectory tracking for high speed flight inside cluttered indoor environments using compensation for aerodynamic drag\remove{in their controller implementation}. \citep{zhang2019maximum} reach flight speed of upto $10$ m/s using a trajectory library generated offline and evaluated online to check for collisions, and performing probability propagation to determine the path reaching the goal position. Focusing on agility of navigation,~\citep{morrell2018differential} exploit a significant body of literature for agile MAVs and demonstrate the benefits of differential flatness transformations for aggressive maneuvers incorporating inverted flight.}

\noindent\textbf{Unconventional Applications of Aerial Robots:} \edit{MAVs have pioneered their way to be integrated in a multitude of application domains including industrial inspection and search and rescue. The AeroARMS Project \citep{ollero2018aeroarms} was aimed at developing robot designs for grabbing and flying with payload, performing inspection tasks based on vision and control feedback. \citep{ollero2021past} discuss the current status and future research for aerial manipulation. The MultiDrone Project~\citep{alcantara2021multidrone} aimed to use multiple autonomous MAVs for media production, demonstrating the use of aerial robots to perform coordinated camera motion. In addition, ~\citep{bonatti2020autonomous} develops an approach for a single robot to execute artistically selected, occlusion free shots. \citep{kamel2018voliro} developed an omnidirectional MAV with recent activities focusing on industrial contact-based inspection.}

\section{Future Directions for Research}
\label{sec:concl}

Concluding this chapter, we proceed to identify \edit{selected open research problems} prohibiting the wider and increasingly effective \edit{and resilient} field deployment of aerial robotics. For robot embodiment, persistent limitations are related to endurance\edit{, safe contact,} and robust physical interaction, without the need for \remove{highly specialized}\edit{heavy} designs that degrade flight performance. Control for \remove{robust}\edit{high performance} free-flight is largely achieved, but \remove{lack the ability of essential physical interaction in conventional designs}\edit{robust and seamless physical interaction is far from established}. Perception and planning are two large open research areas for autonomy. \edit{While significant progress has been made to allow the robots to estimate their state in environments with dust, or obscurants, true resilience against perceptual degradation is yet to be mastered. With robots venturing into dynamic settings, and operating in environments with thin objects, robust perception remains a key challenge}. \edit{Simultaneously, the use of dense map-based representations prevents scalability and accordingly selectively sparse, map-free representations or other means are possibly needed for long-term navigation}. Even with an accurate and dense map reconstruction, challenges in path planning \remove{present other}\edit{lead to} limitations. Three major constraints relate to the computational costs of planning in complex \edit{large-scale but confined} worlds \edit{as highlighted by the DARPA Subterranean Challenge}, limited robustness \remove{observed in}\edit{during} collaborative planning, and the need for more \edit{low-latency} \remove{direct }navigation policies \edit{directly} from sensor data in dynamic and degraded scenes. The latter can possibly be addressed with the emergence of learning-based methods \edit{as demonstrated in attractive domains such as drone racing}. Aerial field robotics is a growing field, and \edit{the} innovative work from individual research teams\edit{, cross-fertilization between industry and academia,} and ambitious robotics competitions \edit{will continue to set milestones in the domain. The research efforts for resilient autonomy will be defining towards rendering aerial robots widely and reliably utilized in a host of diverse and challenging applications in complex industrial and natural environments. In this context, focused research on perception, planning, control and learning, as well as efforts for a unified perception-action loop are particularly important.} 



\bibliographystyle{spbasic}  
\bibliography{bib/references} 


\end{document}